\documentclass[conference]{IEEEtran}
\IEEEoverridecommandlockouts
\usepackage{cite}
\usepackage{booktabs}
\usepackage{subcaption}
\usepackage[dvipsnames, table]{xcolor}
\usepackage[hidelinks, colorlinks=true, citecolor=orange]{hyperref}
\usepackage{amsmath,amssymb,amsfonts}
\usepackage{algorithmic}
\usepackage{graphicx}
\usepackage{textcomp}
\usepackage{multirow}
\usepackage{multicol}
\usepackage{url}
\usepackage{wrapfig}
\usepackage{amsfonts}       
\usepackage{siunitx}
\usepackage{physics}
\usepackage{algorithm}
\usepackage{algorithmic}
\usepackage[font=small]{caption}
\usepackage{tcolorbox}
\usepackage{listings} 
\tcbuselibrary{listings,breakable,skins}

\newcommand{\legenditem}[2]{%
  \fcolorbox{black}{#1!20}{\rule{1em}{1em}}~#2
}
\newtcolorbox[auto counter, number within=section]{promptbox}[2][]{
    enhanced,
    breakable,
    colback=gray!10,
    colframe=black,
    fontupper=\small,
    title={Prompt \thetcbcounter: #2},
    label={#1},
    parbox=false,
    grow to right by=0mm,
    grow to left by=0mm,
    top=2pt,
    bottom=2pt,
    before skip=0pt,
    after skip=0pt,
}

\newtcolorbox{highlightblock}[1][]{%
    enhanced jigsaw, 
    breakable,
    colback=#1!20,
    colframe=#1!60!black,
    boxrule=0pt,
    leftrule=0pt,
    rightrule=0pt,
    toprule=0pt,
    bottomrule=0pt,
    sharp corners,
    parbox=false,
    before skip=2pt,
    after skip=2pt,
}

\newcommand{\fS}{\mathcal{S}}
\newcommand{\fA}{\mathcal{A}}

\newcommand{\R}[1][]{\mathbb{R}^{#1}}

\newcommand{\E}{\mathbb{E}}

\begin{document}

\title{Prompt-Driven Domain Adaptation for End-to-End Autonomous Driving via In-Context RL}

\author{
\IEEEauthorblockN{Aleesha Khurram, Amir Moeini, Shangtong Zhang, Rohan Chandra}
\IEEEauthorblockA{{Dept. of Computer Science, University of Virginia}\\\texttt{\{wma9tt, amoeini, shangtong, rohanchandra\}@virginia.edu}}
}

\maketitle

\begin{abstract}

Despite significant progress and advances in autonomous driving, many end-to-end systems still struggle with domain adaptation (DA), such as transferring a policy trained under clear weather to adverse weather conditions. Typical DA strategies in the literature include collecting additional data in the target domain or re-training the model, or both. Both these strategies quickly become impractical as we increase scale and complexity of driving. These limitations have encouraged investigation into few-shot and zero-shot prompt-driven DA at inference time involving LLMs and VLMs. These methods work by adding a few state-action trajectories during inference to the prompt (similar to in-context learning). However, there are two limitations of such an approach: $(i)$ prompt-driven DA methods are currently restricted to perception tasks such as detection and segmentation and $(ii)$ they require expert few-shot data. In this work, we present a new approach to inference-time few-shot prompt-driven DA for closed-loop autonomous driving in adverse weather condition using in-context reinforcement learning (ICRL). Similar to other prompt-driven DA methods, our approach does not require any updates to the model parameters nor does it require additional data collection in adversarial weather regime. Furthermore, our approach advances the state-of-the-art in prompt-driven DA by extending to closed driving using general trajectories observed during inference. Our experiments using the CARLA simulator show that ICRL results in safer, more efficient, and more comfortable driving policies in the target domain compared to state-of-the-art prompt-driven DA baselines.

\end{abstract}

\section{Introduction}

Autonomous driving systems have achieved remarkable reliability under ideal conditions (such as clear weather or sparse traffic), but their performance degrade sharply in hazardous weather such as heavy rain, snow, and fog or in dense traffic~\cite{chandra2022towards, chandra2022game, chandra2022gameplan, suriyarachchi2022gameopt, suriyarachchi2024gameopt+}. Adverse weather reduces visibility, corrupts LiDAR data, and introduces artifacts that impair camera-based perception~\cite{kothandaraman2021ss, kothandaraman2021bomudanet}. These effects create a distribution shift between training (often clear weather or sparse traffic) and deployment (adverse weather or dense traffic) domains, leading to safety risks and degraded decision-making~\cite{Hegde2023SFUDA3D}. Since most deep learning-based perception models are trained on ideal conditions, they inherit biases in those conditions and lack the robustness needed for target domains~\cite{chandra2019traphic, chandra2019forecasting, chandra2019robusttp}. Moreover, traditional offline models cannot adapt to unforeseen conditions without extensive retraining, which is costly and impractical~\cite{chandra2019densepeds, chandra2020roadtrack, mavrogiannis2022b, wu2023intent}. This motivates the need for agents that can adapt dynamically to shifting driving domains.  

Recent advances in domain adaptation (DA) have shown promise for perception robustness under different conditions. In particular, unsupervised domain adaptation (UDA) methods aim to adapt models trained on labeled clear-weather data to unlabeled adverse-weather domains \cite{Hegde2023SFUDA3D,Zhao2024UniMix}. However, UDA in practice is challenging: collecting labeled data across the full spectrum of driving conditions is prohibitively expensive, fine-tuning large perception models is computationally costly, and re-training for every new environment is impractical for real-world deployment. 

These difficulties motivate a shift toward zero-shot and few-shot approaches~\cite{song_reward_2025}, which seek to bypass labeled target data altogether by exploiting alternative sources of supervision, such as prompts, language descriptions, or synthetic augmentations \cite{Fahes2023PODA,Yang2024ULDA,Marathe2023WEDGE,Park2024RethinkAugLiDAR}. For example, prompt-driven few-shot DA allows pre-trained models to adapt to unseen domains based on textual descriptions alone, while physics-inspired LiDAR augmentations mimic scattering and attenuation effects. Prompt-driven DA works by generating a textual representation of a scene, feeding it into an LLM, and using the LLM's output to guide an agent's action policy. However, prompt-driven DA has only been successful in perception. On the other hand, planning, control, and closed-loop navigation remain reliant on heuristics (e.g., increased following distances), sensor fusion (e.g., radar to compensate for fog-impaired LiDAR), or anticipative localization~\cite{Tan2023RainLoc,Sacoransky2023Radar,Chae2024CVPR,Chae2024ECCV}. To date, prompt-driven adaptation has not extended to closed-loop navigation systems, leaving a critical gap between perception resilience and safe decision-making in hazardous environments.  
\begin{figure}[t]
    \centering
    
    \begin{subfigure}{0.492\linewidth}
        \centering
        \includegraphics[width=\linewidth]{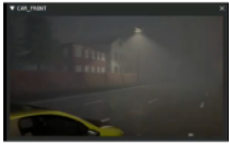} 
        \caption{CoT (front view)}
    \end{subfigure}
    \hfill
    \begin{subfigure}{0.492\linewidth}
        \centering
        \includegraphics[width=\linewidth]{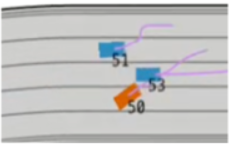} 
        \caption{CoT (top view)}
    \end{subfigure}
    
    \vskip\baselineskip
    \begin{subfigure}{0.492\linewidth}
        \centering
        \includegraphics[width=\linewidth]{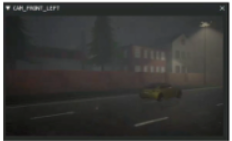} 
        \caption{ICRL (front view)}
    \end{subfigure}
    \hfill
    \begin{subfigure}{0.492\linewidth}
        \centering
        \includegraphics[width=\linewidth]{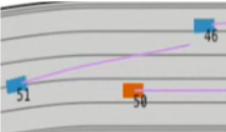} 
        \caption{ICRL (top view)}
    \end{subfigure}
    
    \caption{An example of using prompt-driven domain adaptation (DA) to enable an ego-agent (\textcolor{orange}{\textbf{orange}}) to change lanes in inclement weather in dense traffic (\textcolor{RoyalBlue}{\textbf{blue}} vehicles). \textit{(top row)} Current prompt-driven DA baselines such as Chain-of-Thought (CoT)~\cite{wei2022chain} fail to safely change lanes. \textit{(bottom row)} Using ICRL, a driving policy trained in clear weather successfully adapts and safely changes lanes in inclement weather.}
    \label{fig:cot_icrl}
\end{figure}

Few-shot prompt-driven reasoning methods explored in the Natural Language Processing literature~\cite{mahadevan2025gamechat}, such as in-context learning (ICL)~\cite{ICL, self-refine} typically require carefully structured examples (or context) and expert scoring mechanisms, which limit their applicability to real-world, dynamic domains like driving. In contrast, In-Context Reinforcement Learning (ICRL)~\cite{moeini2025survey} generalizes beyond these constraints. Rather than relying on curated exemplars or specialized reward shaping, ICRL can operate with arbitrary contextual inputs and arbitrary reward functions. This generality allows ICRL to adapt flexibly to novel conditions by using feedback signals observed in the current environment, making it better suited for closed-loop tasks such as autonomous driving in hazardous weather.

ICRL prompts a model with sequences that simulate reinforcement learning episodes, including actions, outcomes, and rewards. Unlike zero-shot methods that rely solely on pretraining, ICRL allows the model to refine its policy across rounds by conditioning on in-context reward signals, without any parameter updates.
This creates a closed-loop adaptation mechanism: the agent can ``learn'' from feedback on the fly, correcting mistakes and incorporating preferences during a single deployment episode. However, while ICRL has been explored in controlled demonstrations and toy problems~\cite{song_reward_2025}, its practical utility for safety-critical, high-dimensional tasks like autonomous driving has not been studied.  

\noindent\textbf{Main Contributions:} In this paper, we present the first system that deploys ICRL on a complex practical task such as closed-loop autonomous driving in hazardous weather and dense traffic conditions. By embedding ICRL into the LimSim co-simulation framework~\cite{fu2024limsim++}, we enable a driving agent to dynamically adapt its high-level decision-making policy. Essentially, given a driving agent that has only learned to drive in clear weather and sparse traffic, we use ICRL to adapt the agent to adverse weather and dense traffic using only a few trajectories seen during inference without curating or processing those trajectories. This addresses the missing link in current literature: extending few-shot DA beyond perception into planning and control, where robustness is most critical.  

Our contributions are as follows. We present a new system built upon SUMO~\cite{SUMO2018} for scenario generation and CARLA~\cite{Dosovitskiy17} for photo-realistic and physically-accurate rendering, which integrates ICRL for prompt-driven few-shot DA. Our system enables DA across diverse weather conditions (from clear to moderately inclement to severely inclement) as well as varying traffic densities ($1\times, 2\times, 3\times$). Across these diversities, our system enables safer, more efficient, and more comfortable driving.

\section{Related Work}

\subsection{Unsupervised Domain Adaption in Autonomous Driving}
Prior work~\cite{Zhao2024UniMix} proposes domain-adaptive detection with both image-level and object-level alignment to minimize style and appearance differences between clear and foggy or rainy images, along with adversarial hard example mining and an auxiliary synthetic weather domain to regularize adaptation. Similarly, for LiDAR-based 3D object detection, Hegde et al.\cite{Hegde2023SFUDA3D} demonstrate that inclement weather introduces artifacts like scattered or missing LiDAR points that degrade detection accuracy. They develop a source-free unsupervised domain adaptation technique to adapt a 3D detector to adverse weather without requiring access to source data during adaptation, using strategies like pseudo-labeling on target data. Another line of adaptation work focuses on video and temporal cues. Yang et al.~\cite{Yang2024ULDA} use teacher–student models to exploit temporal information from consecutive frames in adverse weather, plus temporal weather degradation augmentation, to improve video semantic segmentation under heavy weather. Several behavior modeling approaches also generalize across different traffic domains~\cite{chandra2020cmetric, chandra2020graphrqi, chandra2021using}.

\subsection{Zero-Shot and Few-Shot Domain Adaptation in Autonomous Driving}
Fahes et al.~\cite{Fahes2023PODA} introduce PØDA, which adapts a segmentation model trained on normal weather to an unseen adverse condition using only a natural-language description of that condition. By leveraging vision-language embeddings, they learn to transform the source features toward the target domain described by a prompt, for example ``driving through fire'' or ``sandstorm'', without ever seeing actual fire or sandstorm images. Building on this idea, Yang et al.~\cite{Yang2024ULDA} generated descriptions of rare conditions like sandstorms via a large language model and incorporated them into a unified zero-shot domain adaptation benchmark.

\subsection{Autonomous Driving in Hazardous Weather}

While domain adaptation tries to bridge training-testing domain gaps, other research seeks to directly improve perception models’ robustness via data, architecture, or sensor-level innovations. Below we categorize major trends.

\noindent {Multi-Sensor Fusion and Alternate Modalities:} Chae et al.~\cite{Chae2024CVPR} study LiDAR and 4D radar fusion for robust 3D object detection in various weather conditions. Their approach includes feature-level fusion and weather robustness analysis, and demonstrates consistent gains on multi-weather benchmarks. Chae et al.~\cite{Chae2024ECCV} propose a prompting and distillation framework where a pre-trained LiDAR detector is prompted with 4D radar features during training, and then a LiDAR-only student network distills the weather-robust knowledge. Yu et al.~\cite{Yu2023IR2Vis} tackle this via unpaired infrared-to-visible translation for videos, enhancing fine details in IR sequences.

\noindent{Data Augmentation and Simulation:} Park et al.~\cite{Park2024RethinkAugLiDAR} study LiDAR semantic segmentation and analyze how weather corrupts LiDAR scans. In the image domain, synthetic data generation is also effective. Marathe et al.~\cite{Marathe2023WEDGE} create WEDGE, a dataset of images depicting many types of extreme weather generated by a vision-language model. Fine-tuning detectors on WEDGE leads to gains on real-world weather data, such as improved detection of rarely-seen classes on the DAWN benchmark.

\noindent{Architectural and Algorithmic Innovations:} Kalwar et al.~\cite{Kalwar2023GDIP} integrate Gated Differentiable Image Processing layers into the detection network to dynamically filter adverse effects from camera images before detection. By learning filtering parameters within the network, GDIP adapts to visibility conditions and improves detection in rain or low light. 

\noindent{Planning and Navigation in Adverse Weather:} Adaptive cruise control and braking strategies increase following distances and reduce target speeds when rain or fog is detected, to compensate for longer stopping distances~\cite{RoboDrive2024}. Another challenge is localization in weather. Rain or snow can corrupt LiDAR scans and camera images used for localization against maps. Tan and Meghjani~\cite{Tan2023RainLoc} tackle this with an anticipative localization approach in heavy rain. For path planning, Sacoransky et al.~\cite{Sacoransky2023Radar} demonstrate a radar-based road-following method that uses millimetre-wave radar to detect roadside retro-reflectors in low-visibility conditions.

\subsection{LLMs and VLMs in Autonomous Driving}
Comprehensive overviews of VLM use in autonomous driving highlight their roles across perception, navigation, decision-making, end-to-end architectures, and data augmentation~\cite{Zhou2024SurveyVLM, moeini2025survey}. A series of works propose embedding language reasoning into driving models via VLM guidance. {SimpleLLM4AD} structures driving into Graph VQA stages, perception, prediction, planning, and control, where each is addressed through visual question-answering using VLMs~\cite{Zheng2024SimpleLLM4AD}. {VLM-AD} goes further by using VLMs to provide structured reasoning supervision during training of end-to-end planners, without requiring VLMs at inference~\cite{Xu2024VLMAD}. {DriveVLM} integrates VLM-based scene description and hierarchical planning, and deploys a hybrid VLM/traditional stack in real vehicles~\cite{Tian2024DriveVLM}. {DiMA} introduces a framework where a multi-modal language model provides richer planning representations via distillation into a vision-based planner~\cite{Hegde2025DiMA}. Some works cast LLMs themselves into planning roles. For instance, models like DriveGPT4-V2 harness LLM reasoning to guide control via online imitation learning in closed-loop driving setups \cite{Xu2025DriveGPT4V2}.

\section{Background for ICRL}

\textbf{Reinforcement Learning.} 
We model the tasks using Markov Decision Processes (MDPs), consisting of a state space $\fS$, an action space $\fA$, a reward function $r: \fS \to \R$, an initial distribution $p_0 \in \Delta(\fS)$ with $\Delta(\fS)$ denoting the set of probability distributions over $\fS$, and a transition function $p: \fS \times \fA \to \Delta(\fS)$.
At time step \(0\), the initial state \(S_0\) is drawn from \(p_0\). At each subsequent time step \(t\), the agent observes the current state \(S_t\), samples an action \(A_t \sim \pi(S_t)\), and executes it. The environment then produces the next state \(S_{t+1} \sim p(S_t, A_t)\) and a reward \(R_{t+1} \doteq r(S_{t+1})\). This process repeats until time \(T\), which terminates the episode.
In deep RL ~\cite{mnih2015human,schulman2017proximal}, the policy $\pi$ is parameterized by a neural network with parameter $\theta$. A common objective is to find $\theta$ that maximizes $J(\pi) \doteq \E[\sum_{t=1}^T R_t]$.


\noindent\textbf{In-Context Reinforcement Learning (ICRL) in LLMs.}  
ICRL is an emerging paradigm in both the meta-RL literature as a meta-RL approach that doesn't require parameter updates \cite{beck2023survey,moeini2025survey} and in the NLP literature as a strong inference-time scaling method \cite{song_reward_2025,nieEVOLvEEvaluatingOptimizing2024,krishnamurthy2024can}.
In ICRL, the policy $\pi_\theta$ is conditioned on a \textit{context} $C_t$. The state space includes all the possible contexts. Various constructions exist for $C_t$ \cite{moeini2025survey} but a common one is $C_t = \left((S_i, A_i, R_{i+1}, d_{i+1})\right)_{i=0}^t$ where $d_i$ is indicator of the end of an episode. Note that $C_t$ includes not only the current episode, but previous episodes as well.
Various pretraining techniques can induce ICRL capability in $\pi_\theta$ \cite{moeini2025survey}. Moreover, large language models have been shown to exhibit ICRL behavior on a wide range of bandit problems and MDPs when provided with textual representations of $C_t$ \cite{song_reward_2025,nieEVOLvEEvaluatingOptimizing2024}. 
At test time on a new task, actions are sampled iteratively from $\pi_\theta(S_t, C_t)$, executed in the environment, and used to update $C_t$, while $\theta$ remains fixed. Despite the absence of parameter updates, the quality of $A_t$ typically improves as $C_t$ accumulates more information about the task. This effect is known as in-context policy improvement.
In-context policy improvement persists even for tasks outside the pretraining distribution. For instance, \cite{laskin2023incontext} report improvement on bandit problems where the optimal arms are the opposite of those encountered during pretraining, ruling out the hypothesis that $\pi_\theta$ is simply recalling stored solutions. Combined with empirical evidence of exploration, credit assignment, and exploitation across inference episodes \cite{laskin2023incontext}, and theoretical results showing that transformers can simulate reinforcement learning algorithms in-context \cite{wang2025transformers}, the most consistent explanation is that the forward computation of $\pi_\theta$ itself implements an RL procedure over $C_t$. This dynamic is what is referred to as in-context RL.
Our problem requires processing and reasoning about both images and text, thus we use VLMs and LLMs and show that they similarly possess ICRL capability as expected.

{
}

\section{ICRL in Closed-Loop Autonomous Driving}

\begin{figure}[t]
    \centering
    \includegraphics[width=\columnwidth]{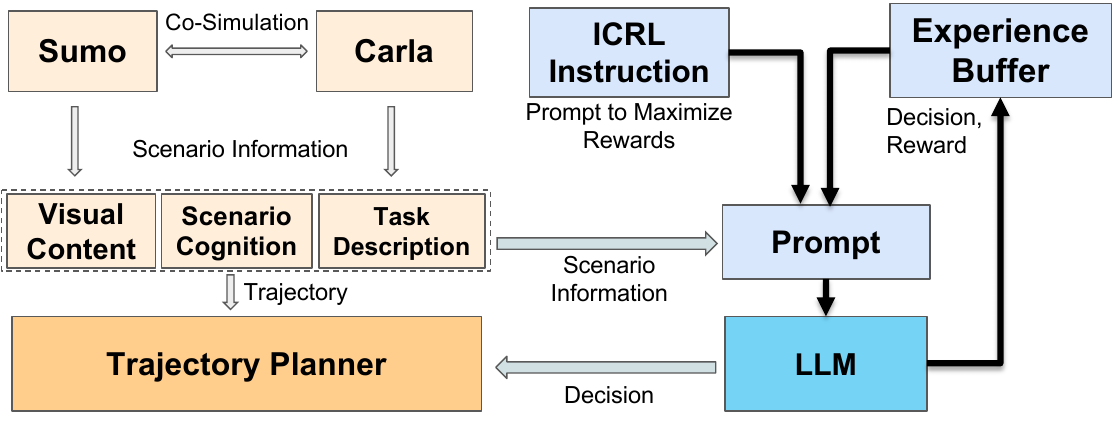}
    \caption{Our system comprises of a driving simulator depicted in \textcolor{YellowOrange}{orange} on the left and ICRL shown in \textcolor{Cerulean}{blue} on the right. The simulator consists of CARLA and SUMO to generate scenes and trajectories. The textual representations of the scene is added to the prompt which then undergoes ICRL to produce a decision that helps adapt the trajectory.}
    \label{fig:approach}
\end{figure}

Our approach is organized into two primary components: the simulation system and the driver agent, with the ICRL algorithm embedded within the latter.
 
\subsection{Driving Simulator}
This module integrates the traffic simulator SUMO~\cite{SUMO2018} and the environment simulator CARLA~\cite{Dosovitskiy17}. SUMO generates realistic traffic scenarios by modeling the underlying road network, defining traffic analysis zones, assigning trips and routes to vehicles, and simulating their movement through the network. These scenarios essentially represent how vehicles flow through an urban environment under different conditions, producing realistic traffic patterns. CARLA, on the other hand, provides sensor outputs and high-fidelity physics-based visual rendering. A bridging module (represented by the dotted boundary) synchronizes the two simulators by keeping SUMO’s traffic logic and CARLA’s visual and sensor world in step with each other. In practice, this means that when a vehicle moves or changes speed in SUMO, the same action is executed in CARLA. The scenario information generated by SUMO and CARLA feeds into a trajectory planner. Given the scenario, the trajectory planner computes an optimal path for the driver agent to execute.


\subsection{Driver Agent}
The driver agent receives multimodal inputs, including panoramic camera images from the ego vehicle and scenario description describing the road network, surrounding vehicles, and navigation goals. This data is then formatted into prompts via the prompt module. Additionally, we supply the driver agent an experience buffer, which contains all of its previous actions and their respective scores. The LLM then generates a driving decision along with an internally assigned numerical reward for that decision. Below, we show an example of the input and output from one of our simulations. This self-assigned reward reflects the LLM’s own evaluation of how well the decision aligns with the performance objectives (safety, comfort, efficiency). Lastly, an evaluation module computes the actual performance metrics from the simulation feedback, which may be used for analysis and reward validation. 

\vspace{5pt}

\begin{promptbox}[lst:llm_prompt]{Demonstrating the Driver Agent}

\begin{center}
    \legenditem{Cerulean}{State} \
\legenditem{YellowOrange}{Context} \
\legenditem{ForestGreen}{Action} \
\legenditem{WildStrawberry}{Output} \
\end{center}

\vspace{3pt}

Input: Current lane description: You are driving on a road with $5$ lanes in your direction, and you are currently driving in the number 4 lane from the left. The length of the current lane is 171.476 m. The limit speed of the current lane is 13.89 m/s. Next lane description: The next lane is too far to consider. \\

\begin{highlightblock}[Cerulean]
Your current state:
Your current position is (675.048, 353.907), speed is 3.993 m/s, acceleration is 1.805 m/s\^{}2, and lane position is 8.828m. \\

\noindent Nearby vehicles description: There are no other vehicles driving near you, so you can drive completely according to your own ideas.\\

\noindent Weather: Note that it is slightly foggy, rainy, and windy. Consequently, the roads are a little wet.\\

\noindent Last decision: The last action you made 1s ago was \texttt{Accelerate}.
\end{highlightblock}

\vspace{7pt}
\begin{highlightblock}[YellowOrange]
\noindent\textbf{Context:}

\noindent Current lane description:  Current lane description: You are driving on a road with $5$ lanes in your direction, and you are currently driving in the number 4 lane from the left. The length of the current lane is 171.476 m. The limit speed of the current lane is 13.89 m/s. Next lane description: The next lane is too far to consider. \\

\noindent Your current state:
Your current position is (675.048, 353.907), speed is 3.993 m/s, acceleration is 1.805 m/s\^{}2, and lane position is 8.828 m. \\

\noindent Nearby vehicles description: There are no other vehicles driving near you, so you can drive completely according to your own ideas. \\

\noindent Weather: Note that it is slightly foggy, rainy, and windy. Consequently, the roads are a little wet. \\

\noindent Last decision: The last action you made 1s ago was \texttt{Accelerate} with a reward of 0.76. \\

\noindent $[\ldots]$
\end{highlightblock}
\vspace{7pt}

\begin{highlightblock}[ForestGreen]
    
Actions: Your available actions are: \\

\noindent\texttt{Accelerate} --- accelerate the vehicle Action\_id: 1 \\

\noindent\texttt{Idle} --- remain in the current lane with current speed Action\_id: 8 \\

\noindent\texttt{Decelerate} --- declerate the vehicle Action\_id: 2 \\

\noindent\texttt{Turn Left} --- change lane to the left of the current lane Action\_id: 3 \\

\noindent\texttt{Turn Right} --- change lane to the right of the current lane Action\_id: 4.
\end{highlightblock}

\vspace{7pt}

Given the current scenario, let's evaluate the best course of action based on safety, efficiency, and comfort, considering the weather conditions and state of the road: \\

\noindent Safety: With the weather being foggy, rainy, and windy, and the roads being a little wet, maintaining a safe speed and control of the vehicle is paramount. Accelerating too quickly could lead to loss of control, especially in these conditions. However, since there are no nearby vehicles, a moderate acceleration to reach a safe and more efficient speed, considering the road conditions, could be considered safe. \\

\noindent Efficiency: Your current speed is significantly below the speed limit of 13.89 m/s. While it's important to drrive safely given the weather conditions, gradually increasing your speed to approach the speed limit, as long as it's safe to do so, would improve efficiency. This would help in reaching your destination quicker without compromising safety. \\

\noindent Comfort: Smooth acceleration, especially in adverse weather conditions, contributes to a more comfortable ride. Abrupt changes in speed or direction could be uncomfortable and potentailly unsafe on wet roads. \\
\begin{highlightblock}[WildStrawberry]

\noindent Output: Considering these factors, the best action would be to continue accelerating moderately. This action balances the need for safety in poor weather conditions, the efficiency of reaching a more appropriate speed for the current road, and the comfort of a smooth ride. \\

\noindent Safety score: 0.8 (due to careful acceleration in adverse conditions) \\

\noindent Efficiency score: 0.7 (as acceleration is needed but must be moderated by conditions) \\

\noindent Comfort score: 0.8 (moderate acceleration is smoother and safer on wet roads) \\

\noindent Response to user: \texttt{Accelerate}\\

\noindent FINAL REWARD: 0.76
\end{highlightblock}

\end{promptbox}

\subsection{ICRL Algorithm} 


%
We use the standard framework introduced in \cite{song_reward_2025} to prompt VLMs and LLMs for ICRL. In this framework, our driving task is modeled with an MDP, and we consider a VLM or an LLM with parameters $\theta$ to be the policy $\pi_\theta$. The action space (depicted in \textcolor{ForestGreen}{\textbf{green}} in the prompt example) for this policy is a finite, discrete set where $A_i \in$ \{{Accelerate}, {Decelerate}, {Idle} {Turn Right}, {Turn Left}\}, which is parsed to get the corresponding action from the trajectory planner. The state space $\fS$ (depicted in \textcolor{Cerulean}{\textbf{blue}} in the prompt example) comprises of the current position, speed, acceleration, and lane position of the ego-vehicle, along with a description of the surrounding vehicles' states, and a textual description of the weather.

Our reward function evaluates a trajectory according to safety, comfort, and efficiency~\cite{fu2024limsim++}. Safety measures collision avoidance and is measured as follows. 
\begin{equation}
    r_{s} = \begin{cases}
        1, & \text{ } \tau_{e} \geq \tau_{threshold} \\
        \tau_{e}/\tau_{threshold}, & else
    \end{cases}
\end{equation}
where $\tau_{e}$ represents the Time to Conflict (TTC) of the agent vehicle, and $\tau_{threshold}$ represents the threshold of TTC. When TTC falls below this threshold value, it indicates the presence of potential risks. Comfort evaluates the smoothness and jerkiness of the driving. One of the most significant factors impacting jerking is lateral and longitudinal acceleration, which must be minimized for a comfortable ride. The calculation of ride comfort is the following. 
\begin{equation}
    r_{c} = \left(s_{x_{a}} + s_{x_{j}} + s_{y_{a}} + s_{y_{j}}\right)/4,
\end{equation}
where $s_{x_{a}}$, $s_{x_{j}}$, $s_{y_{a}}$, and $s_{y_{j}}$ represent lateral acceleration, lateral jerk, longitudinal acceleration, and longitudinal jerk, respectively. Lastly, efficiency evaluates how quickly the agent completes the task, taking into consideration the agent's velocity compared to the velocity limit and the velocity of surrounding vehicles. For maximal efficiency, the agent vehicle's velocity should be as close to the velocity limit as possible. However, in regular and heavy traffic conditions, the vehicle should maintain a velocity at least as much as the average velocity of surrounding vehicles. The efficiency score is given by 
\begin{equation}r_{e} = \begin{cases}
    1, & v_{e} \geq v^{*} \\
    v_{e}/v^{*}, & else
\end{cases}
\end{equation}

\noindent where $v_{e}$ represents the velocity of the ego vehicle, and $v^{*} \in \{v_{avg}, v_{limit}\}$, where $v_{avg}$ and $v_{limit}$ are the average velocity of surrounding vehicles and the velocity limit, respectively. Each individual metric is scored between $0$ and $1$. The reward is sparse such that only the last action of each episode has a non-zero reward. 

The inference process starts with $S_0$, which contains the task description and the initial observation. Within the driver agent, the ICRL algorithm conditions each new decision on both the current scenario information and the full sequence of past decision–reward pairs stored in the experience buffer (depicted in \textcolor{YellowOrange}{\textbf{orange}} in the prompt example). At each time step, scenario information is combined with an ICRL instruction directing the model to maximize cumulative rewards. This prompt is constructed by concatenating the instruction, the scenario state, and the retrieved history from the experience buffer. Then, the LLM outputs a decision and an accompanying self-assigned reward (depicted in \textcolor{WildStrawberry}{\textbf{red}} in the prompt example). This decision is executed in simulation, and the resulting performance feedback is recorded for future steps. This closed-loop design enables adaptation without retraining, as the agent incrementally improves its behavior within a single episode by leveraging sequential reward feedback.

\section{Experiments and Results}

We aim to answer the following research questions:
\textit{RQ1: How does ICRL compare to existing prompt-driven DA techniques in autonomous driving?} We compare ICRL against three prompt-driven baseline methods: Chain-of-Thought (CoT)~\cite{wei2022chain}, Best-of-N~\cite{best-of-n}, and Self-Refine~\cite{self-refine}. We evaluate all methods on three key driving metrics: safety (collision avoidance), comfort (smooth driving without jerky movements), and efficiency (maintaining appropriate speed) in challenging driving scenarios under adverse weather conditions and increasing traffic density. \textit{RQ2: Is prompt-driven DA required?} For this experiment, we conduct an ablation study where we directly task the LLM to generate a response without applying ICRL.


\begin{table*}[htbp]
    \caption{\texttt{Town06}. We compare ICRL with CoT, Best-of-N, and Self-Refine and evaluate safety, comfort, and efficiency in varying traffic densities. Higher is better. Best is \textbf{bolded}. Dark and light shades of blue indicate entries where ICRL is best and second-best, respectively.}
    \centering
    \setlength{\tabcolsep}{6pt}
    \begin{tabular}{rrccccccccc}
    \toprule
    & & \multicolumn{3}{c}{1$\times$ density } & \multicolumn{3}{c}{2$\times$ density } & \multicolumn{3}{c}{3$\times$ density } \\
    \cmidrule(lr){3-5}\cmidrule(lr){6-8}\cmidrule(lr){9-11}
    Weather & Method & Safety & Comfort & Efficiency & Safety & Comfort & Efficiency & Safety & Comfort & Efficiency \\
    \midrule
    \multirow{5}{*}{Clear Weather}
      & CoT~\cite{wei2022chain}          & 0.82 & \textbf{1.00} & 0.82 & 0.02 & 0.60 & 0.78 & 0.08 & \textbf{0.98} & 1.00 \\
      & Best-of-N~\cite{best-of-n}       & 1.00 & 0.76 & 1.00 & 0.99 & 0.86 & 0.71 & 0.81 & 0.86 & 0.80 \\
      & Self-Refine~\cite{self-refine}   & 1.00 & 0.59 & \textbf{1.00} & 1.00 & 0.83 & \textbf{0.99} & 0.00 & 0.77 & 0.89 \\
      & w/o ICRL          & 1.00 & 0.84 & 0.50 & 0.05 & 0.84 & 0.65 & 0.07 & 0.84 & 0.76 \\
      \cmidrule{3-11}& \textbf{ICRL}                    & \cellcolor{Cerulean!80}\textbf{1.00} & 0.65 & 0.88 & \cellcolor{Cerulean!80}\textbf{1.00} & \cellcolor{Cerulean!80}\textbf{1.00} & \cellcolor{Cerulean!20}0.80 & \cellcolor{Cerulean!80}\textbf{1.00} & 0.89 & \cellcolor{Cerulean!80}\textbf{1.00} \\
    \cmidrule{2-11}
    \multirow{5}{*}{Slightly Inclement Weather}
      & CoT~\cite{wei2022chain}          & 1.00 & 0.84 & 1.00 & 1.00 & \textbf{0.99}& 0.73 & 0.12 & 0.88 & 1.00 \\
      & Best-of-N~\cite{best-of-n}       & 1.00 & 0.79 & 1.00 & 1.00 & 0.90 & 0.65 & \textbf{1.00} & \textbf{1.00} & 0.11 \\
      & Self-Refine~\cite{self-refine}   & 1.00 & 0.65 & 1.00 & 1.00 & 0.90 & 0.07 & 0.01 & 0.62 & 1.00 \\
       & w/o ICRL   & 0.99 & 0.86 & 0.79 & 0.56 & 0.84 & 0.07 & 0.02 & 0.84 & 0.76 \\
      \cmidrule{3-11}& \textbf{ICRL}                    & \cellcolor{Cerulean!80}\textbf{1.00} & \cellcolor{Cerulean!80}\textbf{0.85} & \cellcolor{Cerulean!80}\textbf{1.00} & \cellcolor{Cerulean!80}\textbf{1.00} & 0.72 & \cellcolor{Cerulean!80}\textbf{0.93} & \cellcolor{Cerulean!20}0.86 & 0.73 & \cellcolor{Cerulean!80}\textbf{1.00} \\
    \cmidrule{2-11}
    \multirow{5}{*}{Moderately Inclement Weather}
      & CoT~\cite{wei2022chain}          & 1.00 & 0.60 & 0.78 & 1.00 & 0.65 & {1.00} & 0.10 & 0.92 & 1.00 \\
      & Best-of-N~\cite{best-of-n}       & 1.00 & 0.49 & 1.00 & 1.00 & 0.69 & 0.82 & 1.00 & 0.92 & 0.22 \\
      & Self-Refine~\cite{self-refine}   & 1.00 & 0.65 & \textbf{1.00} & 1.00 & 0.45 & \textbf{1.00} & 0.46 & 0.68 & 0.94 \\
      & w/o ICRL   & 0.99 & 0.70 & 0.75 & 0.50 & 0.92 & 0.76 & 0.38 & 0.91 & 0.77 \\
      \cmidrule{3-11}& \textbf{ICRL}                    & \cellcolor{Cerulean!80}\textbf{1.00} & \cellcolor{Cerulean!80}\textbf{0.73} & 0.93 & \cellcolor{Cerulean!80}\textbf{1.00} & \cellcolor{Cerulean!80}\textbf{0.74} & 0.83 & \cellcolor{Cerulean!80}\textbf{1.00} & \cellcolor{Cerulean!80}\textbf{0.95} & \cellcolor{Cerulean!80}\textbf{1.00} \\
    \cmidrule{2-11}
    \multirow{5}{*}{Severely Inclement Weather}
      & CoT~\cite{wei2022chain}          & 0.67 & 1.00 & 1.00 & 1.00 & 0.58 & 1.00 & 0.04 & {1.00} & 1.00 \\
      & Best-of-N~\cite{best-of-n}       & 0.64 & 0.80 & 1.00 & 1.00 & \textbf{0.85} & 0.39 & 1.00 & \textbf{1.00} & 0.72 \\
      & Self-Refine~\cite{self-refine}   & 0.85 & 0.61 & 1.00 & 1.00 & 0.77 & \textbf{1.00} & 1.00 & 0.78 & 1.00 \\
       & w/o ICRL   & 0.98 & 0.93 & 0.76 & 0.54 & 0.93 & 0.77 & 0.37 & 0.93 & 0.78 \\
      \cmidrule{3-11}& \textbf{ICRL}                    & \cellcolor{Cerulean!80}\textbf{1.00} & \cellcolor{Cerulean!80}\textbf{1.00} & \cellcolor{Cerulean!80}\textbf{1.00} & \cellcolor{Cerulean!80}\textbf{1.00} & \cellcolor{Cerulean!20}0.77 & 0.74 & \cellcolor{Cerulean!80}\textbf{1.00} & 0.80 & \cellcolor{Cerulean!80}\textbf{1.00} \\
    \bottomrule
    \end{tabular}
    \label{tab: town6_table}
    \vspace{-5pt}
\end{table*}

\begin{table*}[htbp]
    \caption{\texttt{Town05}. We compare ICRL with CoT, Best-of-N, and Self-Refine and evaluate safety, comfort, and efficiency in varying traffic densities. Higher is better. Best is \textbf{bolded}. Dark and light shades of blue indicate entries where ICRL is best and second-best, respectively.}
    \centering
    \setlength{\tabcolsep}{6pt}
    \begin{tabular}{rrccccccccc}
    \toprule
    & & \multicolumn{3}{c}{1$\times$ density } & \multicolumn{3}{c}{2$\times$ density } & \multicolumn{3}{c}{3$\times$ density } \\
    \cmidrule(lr){3-5}\cmidrule(lr){6-8}\cmidrule(lr){9-11}
    Weather & Method & Safety & Comfort & Efficiency & Safety & Comfort & Efficiency & Safety & Comfort & Efficiency \\
    \midrule
    \multirow{5}{*}{Clear Weather}
      & CoT~\cite{wei2022chain}          & 0.45 & \textbf{1.00} & \textbf{1.00} & 0.15 & \textbf{1.00} & \textbf{1.00} & \textbf{1.00} & \textbf{0.83} & 0.44 \\
      & Best-of-N~\cite{best-of-n}       & 0.96 & 0.84 & 0.71 & 0.80 & 0.81 & 0.84 & 0.72 & 0.78 & 0.74 \\
      & Self-Refine~\cite{self-refine}   & 0.93 & 0.81 & 0.68 & \textbf{0.93} & 0.87 & 0.84 & 0.84 & 0.82 & 0.73 \\
      & w/o ICRL & 0.79 & 0.72 & 0.75 & 0.77 & 0.81 & 0.82 & 0.73 & 0.79 & 0.78 \\
      \cmidrule{3-11}& \textbf{ICRL}                    & \cellcolor{Cerulean!80}\textbf{0.99} & 0.83 & \cellcolor{Cerulean!20}0.83 & \cellcolor{Cerulean!20}0.87 & 0.82 & \cellcolor{Cerulean!20}0.84 & 0.74 & 0.81 & \cellcolor{Cerulean!80}\textbf{0.75} \\
    \cmidrule{2-11}
    \multirow{5}{*}{Slightly Inclement Weather}
      & CoT~\cite{wei2022chain}          & 1.00 & \textbf{0.91} & 0.79 & 0.59 & \textbf{0.90} & 0.84 & 0.12 & 0.88 & \textbf{1.00} \\
      & Best-of-N~\cite{best-of-n}       & 0.93 & 0.81 & 0.72 & 0.81 & 0.81 & 0.84 & 0.77 & 0.82 & 0.83 \\
      & Self-Refine~\cite{self-refine}   & 0.95 & 0.83 & 0.69 & 0.82 & 0.83 & 0.76 & 0.87 & 0.86 & 0.88 \\
      & w/o ICRL & 0.84 & 0.83 & 0.79 & 0.81& 0.78& 0.80 & 0.74 & 0.80 & 0.73 \\
      \cmidrule{3-11}& \textbf{ICRL}                    & \cellcolor{Cerulean!80}\textbf{1.00} & 0.80 & \cellcolor{Cerulean!80}\textbf{0.87} & \cellcolor{Cerulean!80}\textbf{0.87} & \cellcolor{Cerulean!20}0.90 & \cellcolor{Cerulean!80}\textbf{1.00} & \cellcolor{Cerulean!80}\textbf{0.88} & \cellcolor{Cerulean!80}\textbf{0.89} & \cellcolor{Cerulean!20}0.91 \\
    \cmidrule{2-11}
    \multirow{5}{*}{Moderately Inclement Weather}
      & CoT~\cite{wei2022chain}          & {1.00} & \textbf{0.92} & \textbf{0.80} & \textbf{0.95} & \textbf{1.00} & 0.88 & \textbf{0.87} & \textbf{0.96} & 0.80 \\
      & Best-of-N~\cite{best-of-n}       & 0.93 & 0.83 & 0.71 & 0.82 & 0.80 & 0.82 & 0.85 & 0.80 & 0.82 \\
      & Self-Refine~\cite{self-refine}   & 0.94 & 0.81 & 0.69 & 0.90 & 0.85 & 0.87 & 0.75 & 0.77 & 0.79 \\
       & w/o ICRL  & 0.83 & 0.84 & 0.75 & 0.81 & 0.86 & 0.75 & 0.79 & 0.81 & 0.76 \\
      \cmidrule{3-11}& \textbf{ICRL}                    & \cellcolor{Cerulean!80}\textbf{1.00} & 0.80 & 0.68 & \cellcolor{Cerulean!20}0.93 & 0.83 & \cellcolor{Cerulean!80}\textbf{0.89} & 0.83 & \cellcolor{Cerulean!20}0.90 & \cellcolor{Cerulean!80}\textbf{1.00} \\
    \cmidrule{2-11}
    \multirow{5}{*}{Severely Inclement Weather}
      & CoT~\cite{wei2022chain}          & 0.81 & 0.81 & 0.80 & 0.79 & 0.78 & 0.79 & 0.76 & 0.73 & 0.65 \\
      & Best-of-N~\cite{best-of-n}       & 0.92 & \textbf{0.86} & 0.22 & 0.84 & 0.85 & 0.69 & 0.89 & 0.82 & 0.79 \\
      & Self-Refine~\cite{self-refine}   & 1.00 & 0.83 & 0.56 & {0.87} & 0.85 & 0.86 & \textbf{0.90} & 0.89 & \textbf{0.88} \\
       & w/o ICRL    & 0.78 & 0.83 & 0.70 & 0.76  & 0.82 & 0.78 & 0.80 & 0.87 & 0.74\\
      \cmidrule{3-11}& \textbf{ICRL}                    & \cellcolor{Cerulean!80}\textbf{1.00} & 0.74 & \cellcolor{Cerulean!80}\textbf{0.83} & \cellcolor{Cerulean!80}\textbf{0.87} & \cellcolor{Cerulean!80}\textbf{0.88} & \cellcolor{Cerulean!80}\textbf{0.88} & 0.86 & \cellcolor{Cerulean!80}\textbf{0.91} & \cellcolor{Cerulean!20}0.85 \\
    \bottomrule
    \end{tabular}
    \label{tab: town5_table}
    \vspace{-10pt}
\end{table*}

\begin{figure*}[t]
    \centering
    \begin{subfigure}[t]{0.24\textwidth}
        \centering
        \includegraphics[width=\linewidth]{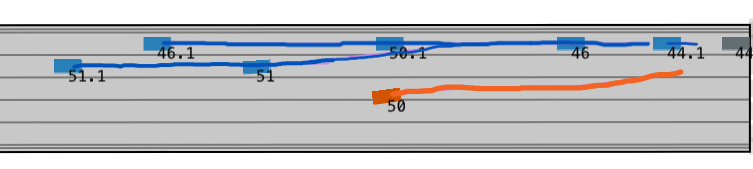}
        \caption{CoT~\cite{wei2022chain} proposes to change to the left lane far too late, causing the agent vehicle to miss the junction.}
        \label{fig:cot_subfig}
    \end{subfigure}
    \begin{subfigure}[t]{0.24\textwidth}
        \centering
        \includegraphics[width=\linewidth]{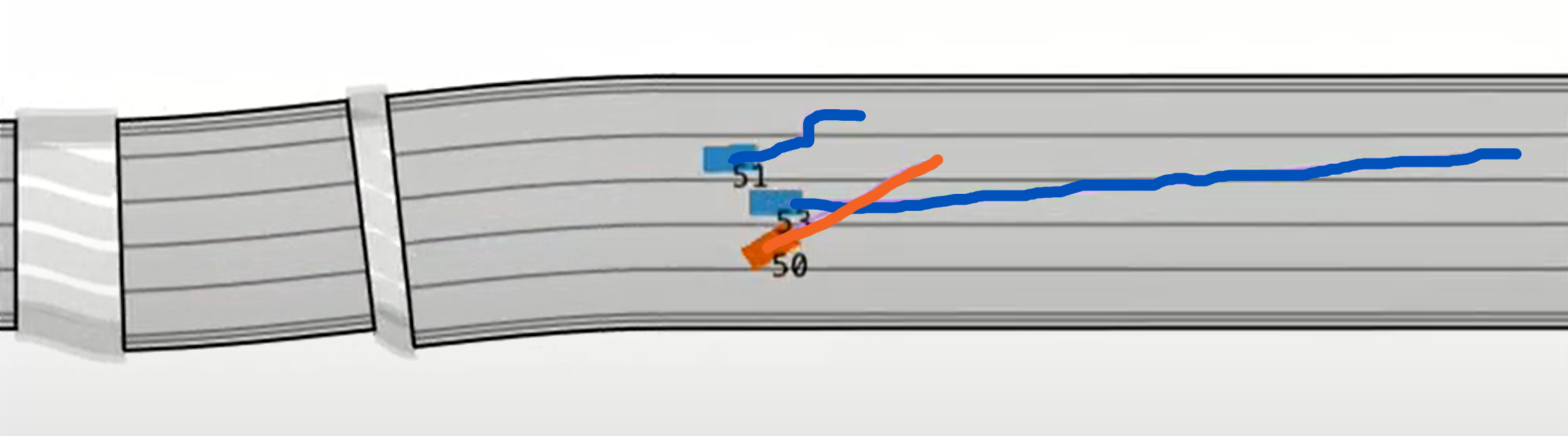}
        \caption{Best-of-N~\cite{best-of-n} proposes a trajectory leading to a near-collision with two vehicles during a lane change}
        \label{fig:Best-of-N_subfig}
    \end{subfigure}
    \begin{subfigure}[t]{0.24\textwidth}
        \centering
        \includegraphics[width=\linewidth]{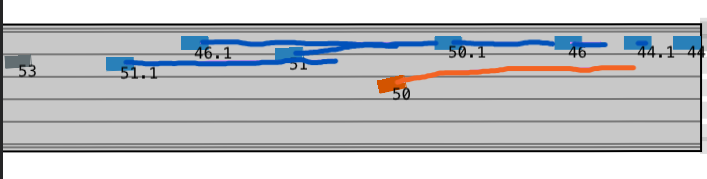}
        \caption{Self-Refine~\cite{self-refine} proposes to change to the left lane too late, causing the agent vehicle to miss the junction.}
        \label{fig:selfrefine_subfig}
    \end{subfigure}
    \begin{subfigure}[t]{0.24\textwidth}
        \centering
        \includegraphics[width=\linewidth]{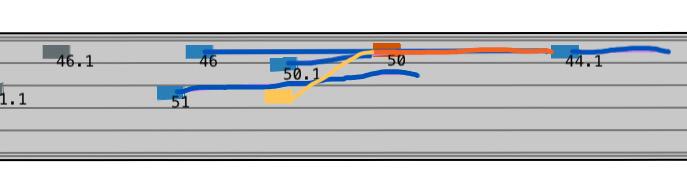}
        \caption{ICRL enables the agent to prioritize shifting lanes, leading it to land the junction correctly.}
        \label{fig:subfig4}
    \end{subfigure}
    \caption{Comparing a lane-changing task using different prompt-driven DA methods. The \textcolor{orange}{\textbf{orange}} vehicle is the ego-agent and the \textcolor{RoyalBlue}{\textbf{blue}} vehicles are the surrounding agents.}
    \label{fig: lane_changes}
\end{figure*}

\begin{figure*}[t]
    \centering
    \begin{subfigure}[t]{0.32\textwidth}
        \centering
        \includegraphics[width=\linewidth]{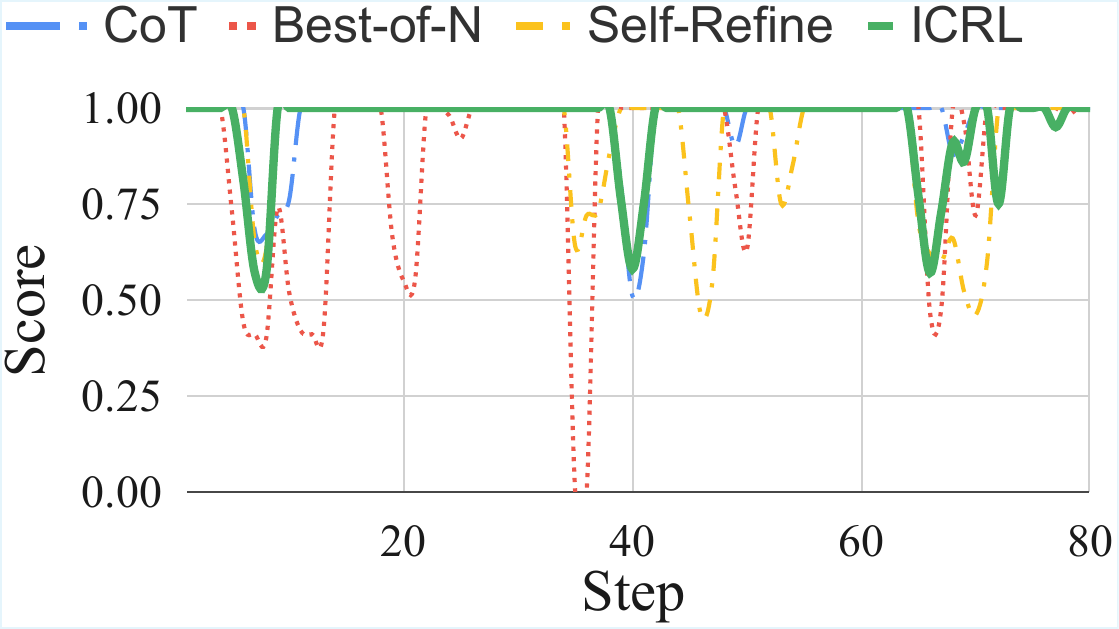}
        \caption{Safety}
        \label{fig:subfig1}
    \end{subfigure}
    \begin{subfigure}[t]{0.32\textwidth}
        \centering
        \includegraphics[width=\linewidth]{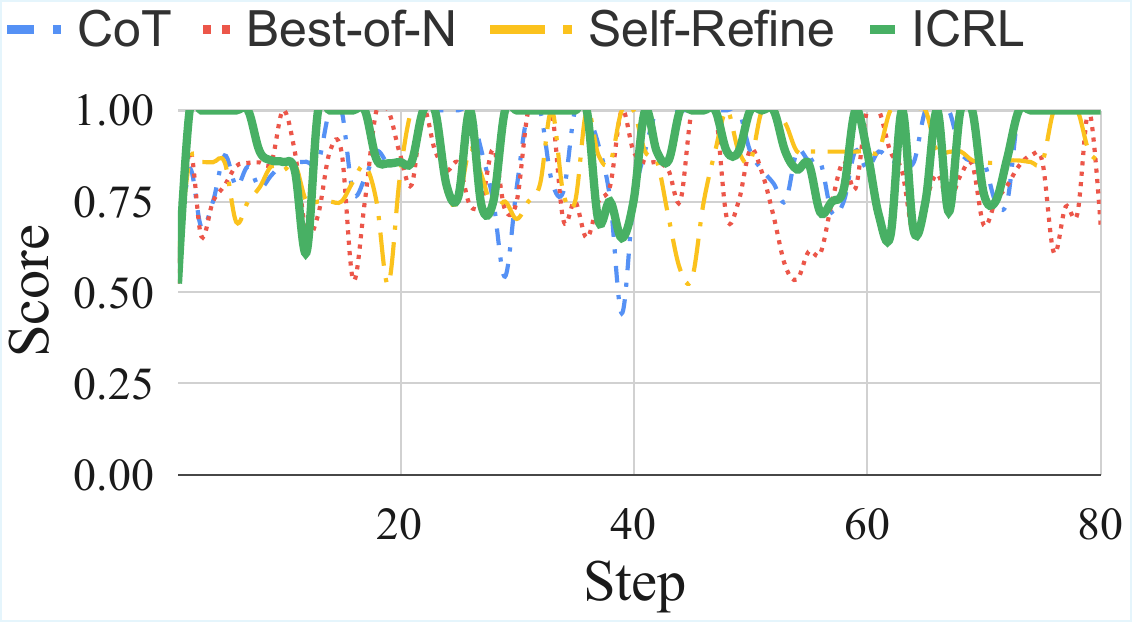}
        \caption{Comfort}
        \label{fig:subfig2}
    \end{subfigure}
    \begin{subfigure}[t]{0.32\textwidth}
        \centering
        \includegraphics[width=\linewidth]{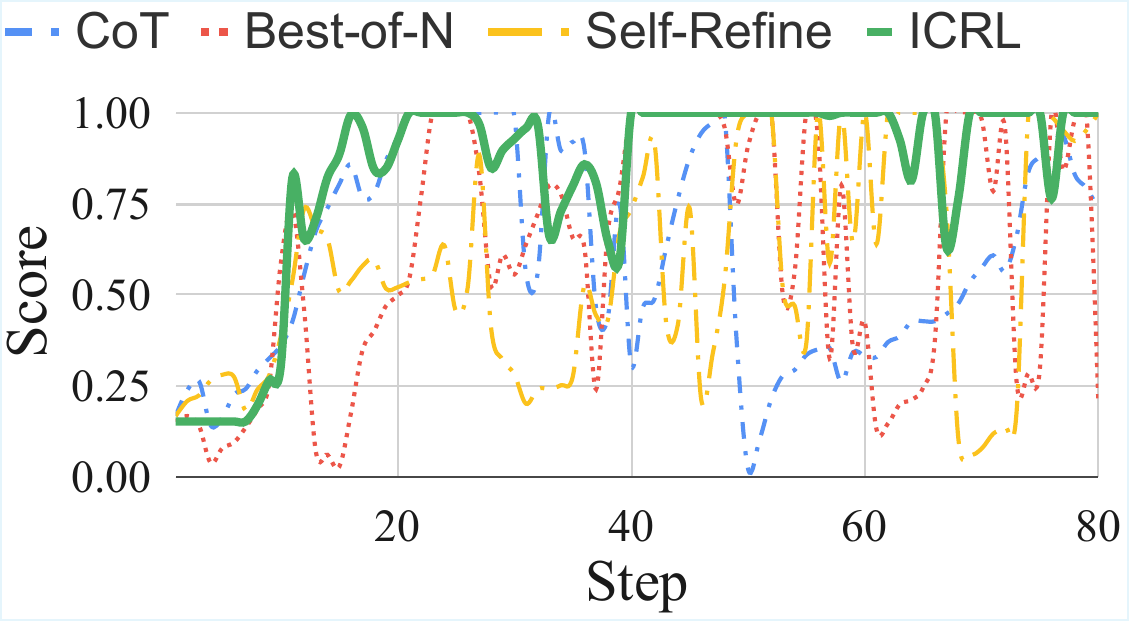}
        \caption{Efficiency}
        \label{fig:subfig3}
    \end{subfigure}
    \caption{We plot the averaged reward metric curves across all episodes. $1$ is the highest, so reward curves that are closer to the top are better.}
    \label{fig:fourfigs}
    \vspace{-10pt}
\end{figure*}

\subsection{Implementation and Scenarios}

Our implementation builds upon the autonomous driving framework, LimSim++, provided by Fu et al.~\cite{fu2024limsim++}. LimSim++ integrates SUMO~\cite{SUMO2018} for traffic simulation with CARLA~\cite{Dosovitskiy17} for photo-realistic and physics-based rendering the environment. We implemented our ICRL approach within the decision making module, replacing the original policy with the inference-time adaptive system that conditions on sequential reward signals and past decisions. This setup let the vehicle agent iteratively refine its driving behaviour dynamically during simulation runs. We conducted experiments on CARLA’s \texttt{Town05} and \texttt{Town06} maps, which feature a four-way intersection and a highway junction as the main navigation challenges, respectively. The agent was prompted every one second in simulation time, and at each step it makes a navigation decision regarding its speed or lane. After the completion of the task, we evaluated the decisions made based on our metrics. We evaluate in the following two maps.

\subsubsection{\texttt{Town06} Map}
CARLA's \texttt{Town06} map features a highway junction. The agent vehicle starts in the center lane of a multilane street and must navigate to the leftmost lane as quickly as possible or else it will miss the junction. In addition to paying attention to other vehicles, it must smoothly turn on the junction and exit on another multilane street.

\subsubsection{\texttt{Town05} Map}
CARLA's \texttt{Town05} map features a four-way intersection, where the agent vehicle must successfully execute a left-turn. Since the agent begins in one of the right lanes, similar to the \texttt{Town06} Map, its priority must be to change lanes as quickly as possible. 

\subsection{Baselines}
Our selected comparative baselines are Chain-of-Thought (CoT)~\cite{wei2022chain}, Best-of-N~\cite{best-of-n}, and Self-Refine~\cite{self-refine}. CoT is one of the most widely used techniques for improving LLM performance at inference time by encouraging step-by-step reasoning. By explicitly decomposing complex problems into intermediate steps, CoT helps models arrive at more accurate and interpretable solutions. Best-of-N is another simple way of optimizing the LLM's output at inference time. At each query, the model generates $N$ independent
samples, evaluates each sample decision with a numerical score, and then selects the highest-scoring candidate. Self-Refine is a multi-step decision-improvement method, wherein the LLM first generates a response to the prompt, and then provides verbal feedback upon which it then improves its response. This process is repeated until some specified stopping condition is reached. In our experiment, our criterion for stopping is convergence on a single navigation decision. For our baselines, we choose $N = 3$. Finally, we conduct an ablation study where we adopt the LLM-drivn driving policy, but by removing the ICRL component, which we call ``w/o ICRL''.

\subsection{ICRL in Adverse Weather}

Tables~\ref{tab: town6_table} and~\ref{tab: town5_table} contain the results. The cells highlighted in dark blue represent the scenarios in which ICRL had the highest score for a particular metric, and the cells highlighted in light blue represent the situations in which ICRL had the second-highest score. There are several interesting observations. First, ICRL's benefits become more evident as the weather becomes more inclement. Specifically, in each table, as we move down the table, the overall density of blue cells increases significantly. The second key observation is that, on average, in inclement weather, ICRL leads in both safety and efficiency (at a marginal cost of comfort) demonstrating an ability to balance both critical objectives, whereas baselines typically tend to sacrifice one or the other. This is also demonstrated in Figure~\ref{fig: lane_changes} where the baselines either fail to change lanes (maximizing safety at the cost of efficiency) or collide during an attempted lane change (maximizing efficiency at the cost of safety). Self-Refine is the only baseline that successfully completed this task, but it noticeably took longer at the junction than ICRL, which featured smooth drivng throughout the task. 

In the \texttt{Town05} map, all baselines have similar scores in safety and comfort, but efficiency of both Best-of-N and Self-Refine are significantly worse. Since this is a common driving scenario, training bias likely leads CoT to outperform the other baselines. However, despite this, ICRL still manages to stay competitive, and outperforms all baselines in safety, which is crucial in inclement weather conditions. 

\subsection{ICRL in Dense Traffic}
In addition to investigating ICRL's performance in inclement weather, we also evaluated its effectiveness in high-density traffic scenarios. 
We tested both $2\times$ and $3\times$ the regular traffic density on both \texttt{Town06} and \texttt{Town05} maps under all weather conditions (Tables~\ref{tab: town6_table} and~\ref{tab: town5_table} and Figure~\ref{fig: lane_changes}). In Figure~\ref{fig:subfig4}, we demonstrate that ICRL enables the agent to merge to the leftmost (or topmost) lane quickly while avoiding multiple nearby vehicles. While some baselines managed partial lane changes, no baseline successfully completed the full merging task due to the increased traffic density. For example, CoT and Self-Refine proposed the lane change far too late (Figures~\ref{fig:cot_subfig} and~\ref{fig:selfrefine_subfig}), prioritizing comfort over efficiency. Best-of-N led the agent to a near collision, sacrificing safety for efficiency (Figure~\ref{fig:Best-of-N_subfig}). Even though ICRL had lower comfort scores on average than other baselines, it prioritized safety and efficiency, leading it to be the only method that consistently completed the task while maintaining safe distances from surrounding vehicles, demonstrating its superior adaptability in complex, multi-constraint environments. Finally, in Figure~\ref{fig:fourfigs}, we qualitatively compare ICRL with the baselines. The higher the curve (and therefore, closer to $1$, the better performing it is). We observe that ICRL is consistently, on average, superior to state-of-the-art prompt-driven DA methods for closed-loop autonomous driving tasks.

\section{Conclusion}

In this work, we presented a new prompt-driven DA approach for closed-loop autonomous driving using ICRL. Using CARLA for traffic simulation, ICRL enabled an agent to dynamically adapt to diverse traffic scenarios such as adverse weather and traffic density on the fly without updating the model parameters. Our experiments demonstrated that ICRL consistently outperforms state of the art prompt-driven DA baselines with respect to safety, comfort, and efficiency. Overall, our work shows that ICRL is a promising strategy to include towards building more adaptive and resilient autonomous driving systems that can handle the unpredictability and diversity in real world traffic conditions.

There are several lines of inquiry that can be investigated following this research. First, we do not have a good understanding of the theoretical aspects of ICRL. For instance there are three key questions: (1) How much contextual information (episodes, reward variance, trajectory diversity) is sufficient for successful adaptation? (2) What structural properties must context prompts satisfy for optimal few-shot behavior? (3) Can ICRL offer performance guarantees, such as regret bounds or convergence properties, under mild assumptions on the environment or reward signal? Second, can ICRL be generalized beyond driving to more real world robot navigation tasks? Here the goal is to embed ICRL as a general layer that sits on top of any navigation stack, thereby enabling off-the-shelf few-shot DA. Finally, in this work, we primarily focused on textual input and LLMs, but future work could use our system to develop more ribust models using multimodal inputs via VLMs and VLA models.

\flushbottom

\bibliography{refs, refs_icrl}
\bibliographystyle{IEEEtran}
\end{document}